\documentclass[11pt,a4paper]{article}

\usepackage[utf8]{inputenc}
\usepackage[T1]{fontenc}
\usepackage{lmodern}
\usepackage[margin=2.5cm]{geometry}
\usepackage{amsmath,amssymb}
\usepackage{graphicx}
\usepackage{booktabs}
\usepackage{siunitx}
\usepackage{tikz}
\usetikzlibrary{positioning, arrows.meta, fit, backgrounds, calc}

\definecolor{chipfill}{RGB}{232,224,245}  
\definecolor{hostfill}{RGB}{224,242,226}  
\definecolor{robotfill}{RGB}{252,236,214} 
\tikzset{
  blk/.style={draw, rounded corners=2pt, fill=white, align=center,
    font=\footnotesize, minimum height=10mm, inner sep=3pt,
    execute at begin node={\hyphenpenalty=10000\relax}},
  arr/.style={-{Stealth[length=2mm]}, thick},
  grp/.style={draw, dashed, rounded corners=4pt, inner sep=6pt},
  grplabel/.style={font=\footnotesize\bfseries, anchor=south west, inner sep=2pt},
  neuron/.style={circle, draw, thick, minimum size=7mm, inner sep=0pt,
    font=\scriptsize},
}
\usepackage{xcolor}
\usepackage{hyperref}
\usepackage[capitalise]{cleveref}

\usepackage{authblk}

\title{Real-time neuromorphic audio and vision perception for autonomous robotic
applications: Integration and experimental evaluation on a rover platform}

\author[]{Guillem Fustier-Castillo, Ruben Ribes-Serrano, Javier Naranjo-Alcazar, Vicent Ortiz-Castello and Pedro Zuccarello}
\affil[]{Instituto Tecnologico de
Informatica (ITI), Valencia, Spain}
\date{}

\begin{document}

\maketitle

\begin{abstract}
Neuromorphic sensing and computing offer a promising alternative to conventional
deep learning pipelines for mobile robots, where perception must run continuously
under tight latency and energy constraints. However, most neuromorphic solutions
are still evaluated offline or on isolated benchmarks, and their integration into
complete robotic systems remains largely unexplored. In this technical report, a
four-wheel rover platform is presented that combines three real-time perception
pipelines: a classical audio tagging pipeline based on Pretrained Audio Neural
Networks (PANNs) running on an NVIDIA Jetson GPU, a neuromorphic audio pipeline in
which a spiking neural network (SNN) is deployed on the SynSense Xylo Audio~2 chip,
and a neuromorphic vision pipeline in which events from an event-based camera are
processed by the SpikeYOLO spiking detector for traffic cone detection. On the Xylo
chip, the whole processing chain---from the microphone signal to the classification
of low- and high-pitched sounds---is performed on-chip, while the host only
configures the device and reads the output neurons. All models are deployed as
containerised services and integrated through ROS~2, and a configurable action
dispatcher maps the detected classes to robot behaviours, closing the loop between
perception and actuation. The platform is demonstrated in a proof-of-concept
scenario in which the robot reacts to sound classes detected by both audio
pipelines and detects traffic cones with the event camera, showing the feasibility
of integrating neuromorphic hardware into a multimodal robotic platform operating in
real time.
\end{abstract}


\section{Introduction}
\label{sec:intro}

Autonomous mobile robots must perceive their environment continuously and react to
it within short time frames, while relying on embedded computing platforms with
limited energy budgets. Deep learning has become the standard approach for robotic
perception, and edge devices such as the NVIDIA Jetson family make it possible to
run convolutional neural networks (CNNs) close to the sensors, reducing the
dependence on network connectivity, the latency of the response and the
transmission of raw data~\cite{andriulo2024edge}. In the audio domain, recent
studies have shown that audio tagging models can be deployed on resource-constrained
devices, although their inference time and thermal behaviour strongly depend on the
selected architecture and the deployment conditions~\cite{bibbo2023audio,
grauharo2025comprehensive}. These constraints are even more critical in battery
powered robots, where perception competes for energy with locomotion and the
remaining onboard systems.

Neuromorphic computing addresses these limitations by taking inspiration from the
way biological neural systems process information. Spiking Neural Networks (SNNs)
process information through discrete spike events instead of continuous-valued
activations, which enables sparse and event-driven computation with high energy
efficiency~\cite{malcolm2023comprehensive, eshraghian2023training}. This paradigm
is complemented by neuromorphic sensors, such as event-based cameras, which
generate spike-like outputs directly~\cite{lichtsteiner2008128}, and by dedicated
neuromorphic processors that execute SNNs at very low power. These properties make
neuromorphic systems particularly attractive for always-on perception in mobile
robotics. Nevertheless, most neuromorphic audio and vision solutions are evaluated
offline or on isolated benchmarks~\cite{baek2024snn, larroza2025spike}, and their
integration into complete robotic platforms operating in closed loop remains
largely unexplored.

In this technical report, the integration of neuromorphic audio and vision
perception into a four-wheel rover platform is presented. The robot combines three
real-time perception pipelines: (i) a classical audio tagging pipeline based on
PANNs~\cite{kong2020panns} running on the GPU of an NVIDIA Jetson device, (ii) a
neuromorphic audio pipeline based on an SNN deployed on the SynSense Xylo Audio~2
chip, in which the audio signal is processed entirely on-chip, and (iii) a
neuromorphic vision pipeline in which the output of an event-based camera is
processed by the SpikeYOLO spiking detector~\cite{luo2024integer} to detect traffic
cones. Running classical and neuromorphic pipelines on the same platform allows
both approaches to be compared under the same operating conditions, while
providing the robot with complementary perception capabilities.

The main contributions of this work are the following:
\begin{itemize}
  \item A multimodal rover platform that integrates conventional and neuromorphic
  sensors and processors, including an NVIDIA Jetson device, a SynSense Xylo
  Audio~2 development kit and an event-based camera.
  \item The closed-loop integration of the Xylo Audio~2 neuromorphic chip into the
  robot, from the deployment of a trained SNN on the hardware to the execution of
  robot actions triggered by its predictions.
  \item An event-based vision pipeline that feeds the SpikeYOLO spiking detector
  with event representations built in real time from the camera stream.
  \item A modular and containerised software architecture, in which each model is
  exposed as an independent service and integrated through ROS~2, so that models
  can be replaced without modifying the robot software.
  \item A proof-of-concept demonstration in which the robot detects sound classes
  with both the classical and the neuromorphic audio pipelines and translates them
  into actions, while detecting traffic cones with the event camera.
\end{itemize}

The remainder of this report is organized as follows. Section~\ref{sec:background}
introduces SNNs, neuromorphic hardware and sensors, and reviews related work.
Section~\ref{sec:platform} describes the robotic platform, and
Section~\ref{sec:architecture} presents the software architecture that integrates
the perception pipelines with the robot. Sections~\ref{sec:panns},
\ref{sec:xylo} and~\ref{sec:vision} detail the classical audio, neuromorphic audio
and neuromorphic vision pipelines, respectively. Section~\ref{sec:evaluation}
describes the proof-of-concept demonstration, Section~\ref{sec:limitations}
discusses the
current limitations and future work, and Section~\ref{sec:conclusion} concludes the
report.

\section{Background and related work}
\label{sec:background}

\subsection{Spiking neural networks and the LIF neuron}

Unlike traditional artificial neural networks (ANNs), which rely on
continuous-valued activations, SNNs process information through discrete spike
events, mimicking the behaviour of biological neurons~\cite{zhang2022recent,
eshraghian2023training}. Each neuron integrates its input over time and emits a
spike only when its internal state exceeds a threshold, so that computation and
communication only take place when relevant information is present. This sparse
and event-driven behaviour is the basis of the energy efficiency of SNNs when they
are executed on dedicated hardware~\cite{malcolm2023comprehensive}.

The most widely used neuron model in SNNs, and the one used by the neuromorphic
audio pipeline of this work, is the Leaky Integrate-and-Fire (LIF) neuron. A LIF
neuron accumulates the incoming spikes, weighted by the synaptic weights, into its
membrane potential, which decays over time with a characteristic time constant.
When the membrane potential reaches a threshold, the neuron emits a spike and its
potential is reset~\cite{eshraghian2023training}.

\subsection{Neuromorphic hardware and sensors}

To harness SNNs in real-world applications, sensor data need to be represented as
spike trains~\cite{auge2021survey}. While conventional sensors require an
additional encoding step, which strongly influences the performance and efficiency
of the network~\cite{larroza2025spike}, neuromorphic sensors are specifically
designed to generate spike-based outputs directly~\cite{lichtsteiner2008128}.

Event-based cameras are the most mature example of neuromorphic sensors. Instead of
capturing frames at a fixed rate, each pixel independently and asynchronously
reports changes in log-intensity as events, each one composed of the pixel
coordinates, a timestamp and a polarity indicating whether the brightness has
increased or decreased~\cite{lichtsteiner2008128, gallego2022event}. As a result,
event cameras offer microsecond temporal resolution, high dynamic range, low power
consumption and no motion blur, which makes them well suited for mobile robots.

On the processing side, several neuromorphic processors have been proposed for the
efficient execution of SNNs, such as TrueNorth~\cite{merolla2014million},
SpiNNaker~\cite{furber2014spinnaker} or Loihi~\cite{davies2018loihi}. In the audio
domain, the SynSense Xylo family offers low-power SNN inference processors for
edge applications~\cite{bos2022sub}. In particular, Xylo Audio~2 combines an analog
audio front-end, which converts the microphone signal into spike trains through a
bank of band-pass filters, with a digital SNN core, enabling end-to-end audio
processing on a single chip. Xylo devices are programmed through the open-source
Rockpool library~\cite{muir2019rockpool}, which allows SNNs to be trained in
PyTorch and mapped to the hardware. The Xylo Audio~2 platform is described in
detail in Section~\ref{sec:xylo-platform}.

\subsection{Related work}

Edge computing for audio tagging has received increasing attention in recent
years, as processing audio close to the sensor reduces the transmission of raw
data, the latency of the response and the dependence on network
connectivity~\cite{andriulo2024edge}. Lamrini et al.~\cite{lamrini2023evaluating}
evaluated pre-trained CNNs for anomaly detection in smart cities on embedded
platforms, and Bibbo et al.~\cite{bibbo2023audio} studied the deployment of a single
CNN-based audio tagging architecture on a Raspberry~Pi, highlighting challenges
related to thermal management and inference latency over extended periods. These
works were extended in~\cite{grauharo2025comprehensive}, where all the PANNs
architectures, together with other CNN-based audio tagging models, were evaluated on
a Raspberry~Pi during continuous 24-hour inference sessions, showing that the
choice of architecture and the deployment conditions determine whether real-time
operation can be sustained. Edge inference nodes of this kind have been integrated
into complete monitoring pipelines with remote supervision through IoT
platforms~\cite{naranjoalcazar2025realtime}, and their deployment in real
industrial environments has shown that AudioSet-trained models may assign high
confidence to wrong classes under real-world conditions~\cite{garciaballesteros2026confidently}.
All these works rely on conventional deep learning models executed on CPUs or GPUs,
and none of them considers a mobile robotic platform.

In the neuromorphic domain, the application of SNNs to sound processing has been
reviewed in~\cite{baek2024snn}, which identifies the need for specialised SNN
architectures tailored to audio tasks. Since conventional microphones do not
produce spikes, a large part of the research has focused on spike encoding
techniques~\cite{auge2021survey}: Speech2Spikes~\cite{stewart2023speech2spikes}
proposed an efficient audio encoding pipeline for real-time neuromorphic systems,
mainly in the context of speech, while~\cite{larroza2025spike} benchmarked several
spike encoders for environmental sound, showing that the choice of the encoder
affects both the classification performance and the spike firing rate, and hence
the energy efficiency, of the network. Dedicated hardware removes this step from
the host: the Xylo family integrates the encoding and the SNN inference on the same
chip, enabling audio processing applications with sub-mW power
consumption~\cite{bos2022sub}. However, these works evaluate neuromorphic audio
processing on datasets or as standalone applications, without integrating it into
a system that acts on its environment.

Regarding vision, event-based cameras have been widely studied for robotics and
computer vision due to their high temporal resolution, high dynamic range and low
power consumption~\cite{lichtsteiner2008128, gallego2022event}. The release of
large-scale event-based detection datasets, such as Gen1 in the automotive
domain~\cite{detournemire2020large}, has fostered the development of object
detectors for event data, and SNN-based detectors have recently narrowed the
performance gap with ANNs. SpikeYOLO~\cite{luo2024integer} combines a simplified
YOLOv8 architecture with integer-valued training and spike-driven inference,
achieving state-of-the-art results among SNNs on COCO and Gen1, while
SpikeDet~\cite{fan2026spikedetbetterfiringpatterns}, the evolution of SpikSSD,
improves the firing patterns of spiking neurons to obtain accurate and
energy-efficient detection. As in the audio case, these detectors are mostly
evaluated offline on benchmark datasets.

Finally, the use of SNNs for robot control has been reviewed
in~\cite{bing2018survey}, and neuromorphic processors such as
Loihi~\cite{davies2018loihi} have been proposed as low-power computing platforms
for embedded and robotic applications. Most of these works, however, focus on a
single sensing modality or on simulated environments.

In contrast to the works above, this report presents a robotic platform in which
neuromorphic audio and vision perception run in real time on the same robot,
together with a classical audio pipeline that serves as a reference. To the best of
our knowledge, this is the first work that integrates the Xylo Audio~2 chip into a
mobile robot in closed loop, so that the predictions computed on-chip directly
trigger robot actions, and that combines it with an event-based spiking detector in
a single modular architecture. 


\section{Robotic platform}
\label{sec:platform}

This section describes the rover used in this work, referred to as LISA. First,
the mechanical design and the kinematic model are presented. Next, the hardware
components and their connections are detailed. Finally, the software stack that
runs on the robot is described.

\subsection{Mechanical design and kinematics}

\begin{figure}[t]
  \centering
  \includegraphics[width=0.75\linewidth]{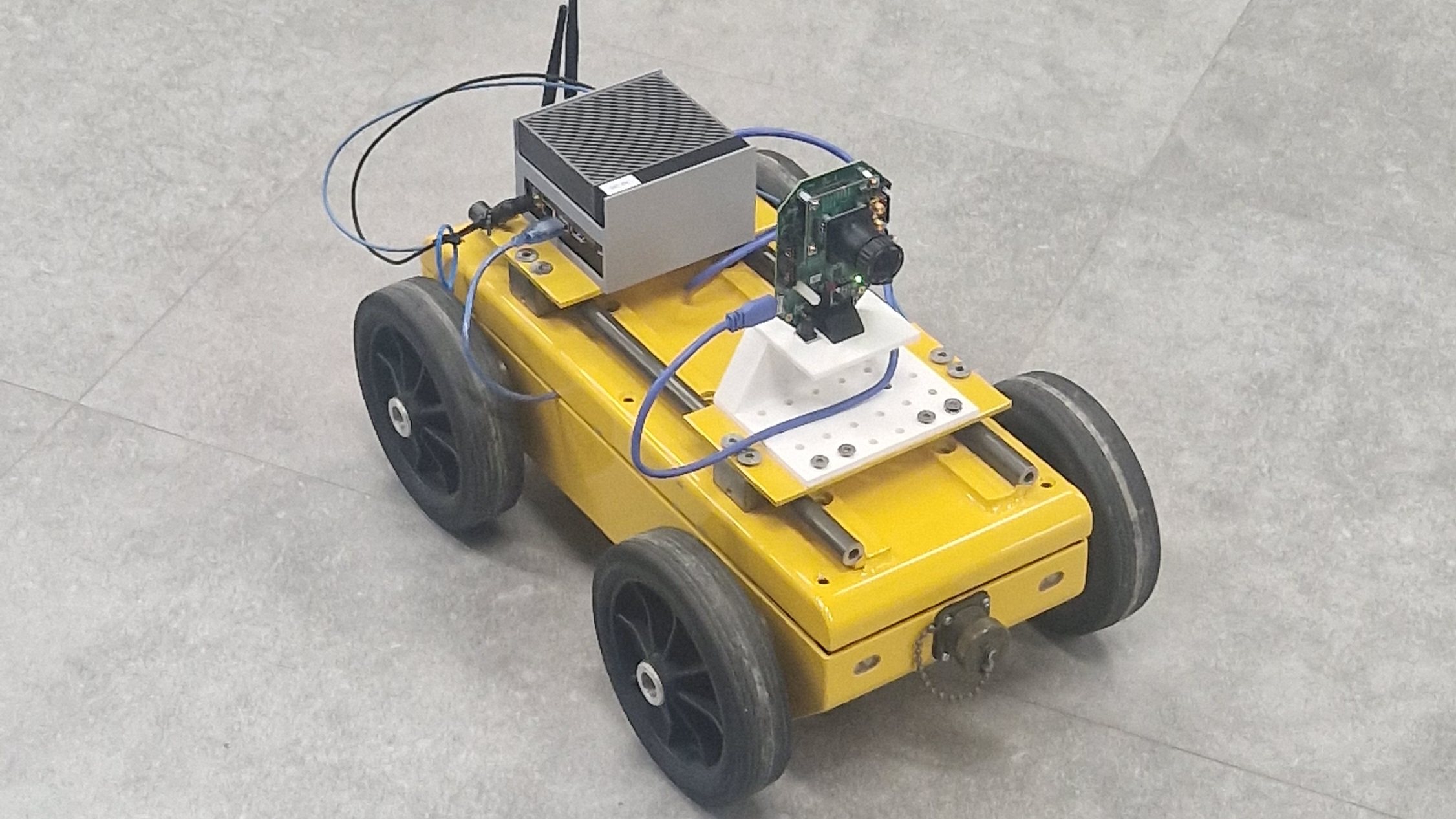}
  \caption{The LISA rover platform, with the Jetson AGX Orin and the event camera mounted on top.}
  \label{fig:robot}
\end{figure}

LISA is a four-wheel skid-steer rover, in which the two wheels on each side are
driven together and the robot turns by commanding different velocities to the left
and right sides (see Fig.~\ref{fig:robot}). The wheels have an effective radius of
\SI{0.077}{m} (\SI{154}{mm} diameter), and each wheel is equipped with an encoder
of 980 pulses per revolution.

\begin{figure}[t]
  \centering
  \includegraphics[width=\linewidth]{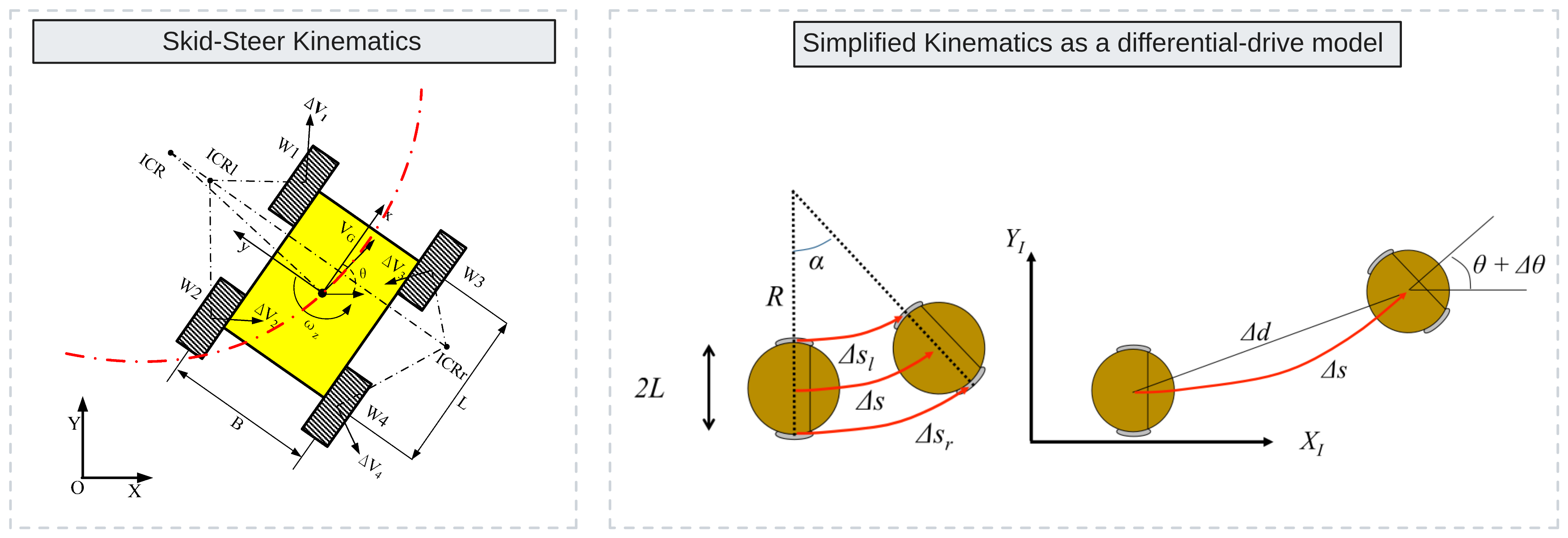}
  \caption{Kinematic model of the rover: skid-steer kinematics (left) and its simplification as a differential-drive model used for wheel odometry (right).}
  \label{fig:kinematics}
\end{figure}

Wheel odometry is computed from the encoder readings using a differential-drive
model, in which the velocities of the left and right wheels are averaged on each
side. Due to the lateral slippage that occurs in skid-steer platforms when turning,
the effective track width differs from the geometric one. The effective track
width was therefore calibrated empirically: during a full turn in place, the
encoders reported approximately \SI{803}{\degree} instead of \SI{360}{\degree},
which corresponds to a scrub factor of 2.23. Consequently, the geometric
half-track of \SI{0.281}{m} is scaled to an effective track width of
\SI{0.627}{m}. The kinematic model of the robot is shown in
Fig.~\ref{fig:kinematics}.

\subsection{Hardware}

\begin{figure}[t]
  \centering
  \includegraphics[width=\linewidth]{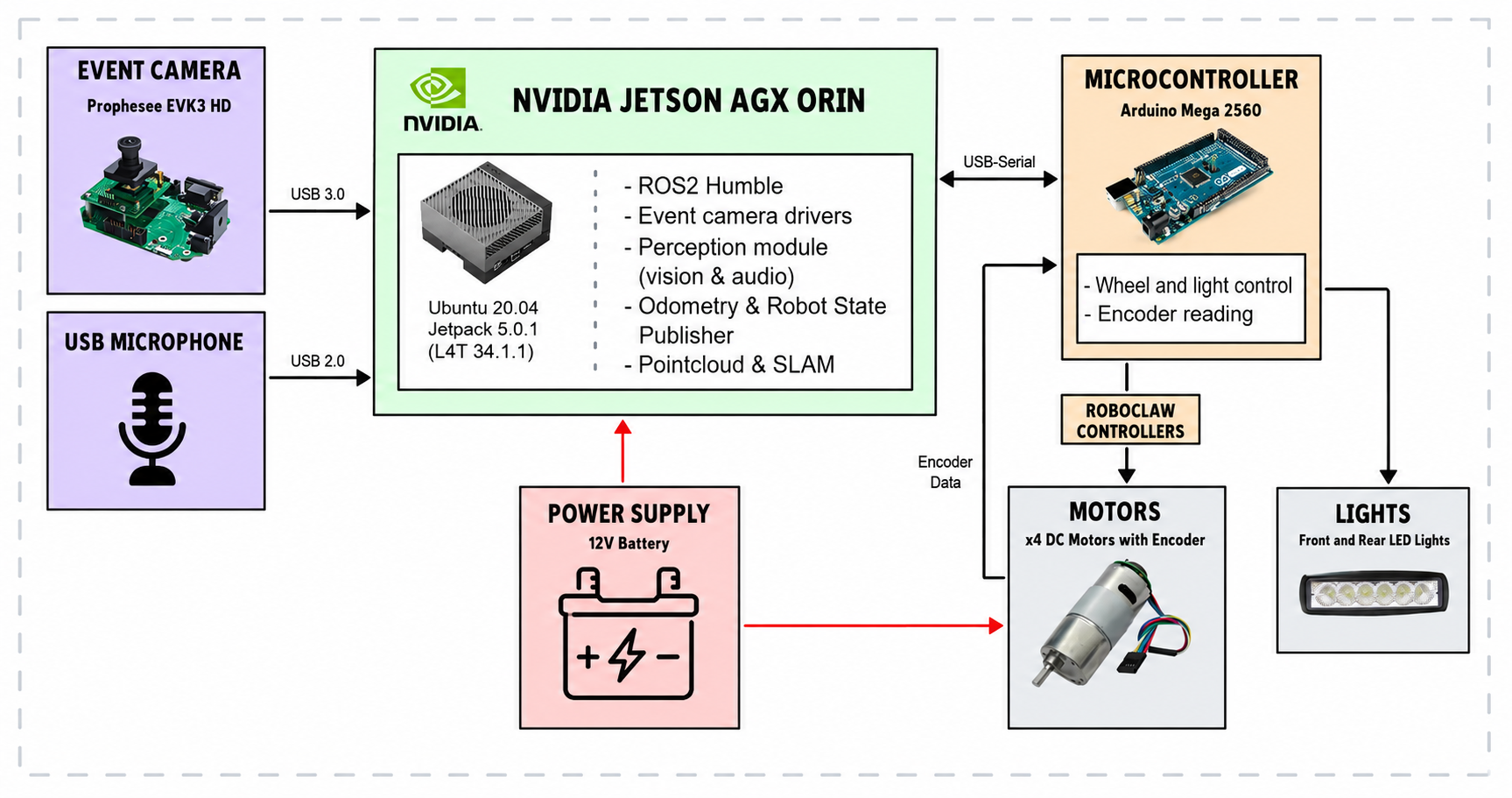}
  \caption{Hardware architecture of the rover.}
  \label{fig:hw-arch}
\end{figure}

The hardware architecture of the robot is shown in Fig.~\ref{fig:hw-arch}, and its
main components are summarised in Table~\ref{tab:hardware}. The robot is composed
of the following components:

\begin{enumerate}
  \item Main computing unit: an NVIDIA Jetson AGX Orin runs the perception models,
  the orchestration software and the ROS~2 nodes of the robot. The system was also
  tested on an NVIDIA Jetson Orin Nano, a lower-cost and lower-power alternative
  (see Section~\ref{sec:panns}).
  \item Neuromorphic audio processor: a SynSense Xylo Audio~2 hardware development
  kit (HDK), connected to the Jetson via USB, which integrates its own microphone
  and executes the neuromorphic audio pipeline.
  \item Event-based camera: a Prophesee EVK3 HD evaluation kit, based on the Sony
  IMX636 event sensor with a resolution of $1280\times720$ pixels, connected via
  USB.
  \item RGB camera: a Logitech C720 HD Webcam, connected via USB and used for visualisation, monitoring and dataset
  recording.
  \item Microphone: a RODE Lavalier II microphone, connected via USB and used by
  the classical audio pipeline.
  \item Low-level controller: an Arduino Mega board, connected to the Jetson
  through a serial link, which forwards the motion commands to the motor drivers,
  controls the front and rear lights and streams the encoder readings back to the
  Jetson.
  \item Motor drivers: two RoboClaw controllers, which drive the motors of the left
  and right sides.
\end{enumerate}

\begin{table}[t]
  \centering
  \caption{Main hardware components of the LISA rover.}
  \label{tab:hardware}
  \begin{tabular}{ll}
    \toprule
    Component & Model \\
    \midrule
    Computing unit & NVIDIA Jetson AGX Orin (also tested on Orin Nano) \\
    Neuromorphic audio processor & SynSense Xylo Audio~2 HDK \\
    Event-based camera & Prophesee EVK3 HD (Sony IMX636, $1280\times720$) \\
    RGB camera & Logitech C720 HD Webcam \\
    Microphone & RODE Lavalier II \\
    Low-level controller & Arduino Mega \\
    Motor drivers & 2 $\times$ RoboClaw \\
    Wheels & Ø \SI{154}{mm}, encoders of 980 pulses/rev \\
    \bottomrule
  \end{tabular}
\end{table}

\subsection{Software stack}

The robot software is built on ROS~2~\cite{macenski2022robot}. The robot nodes
are organised in several packages: a serial bridge that communicates with the
Arduino, converts the velocity commands published on \path{/cmd_vel} and the
light commands published on \path{/led_robot} into serial frames, and integrates
the wheel odometry published on \path{/odom}; a robot description package that
provides the URDF model and the transforms between the robot frames; the camera
drivers, which publish the event stream on \path{/event_camera/events} and the
RGB images on \path{/image_raw}; and the action dispatcher described in
Section~\ref{sec:architecture}.

\begin{figure}[t]
  \centering
  \includegraphics[width=\linewidth]{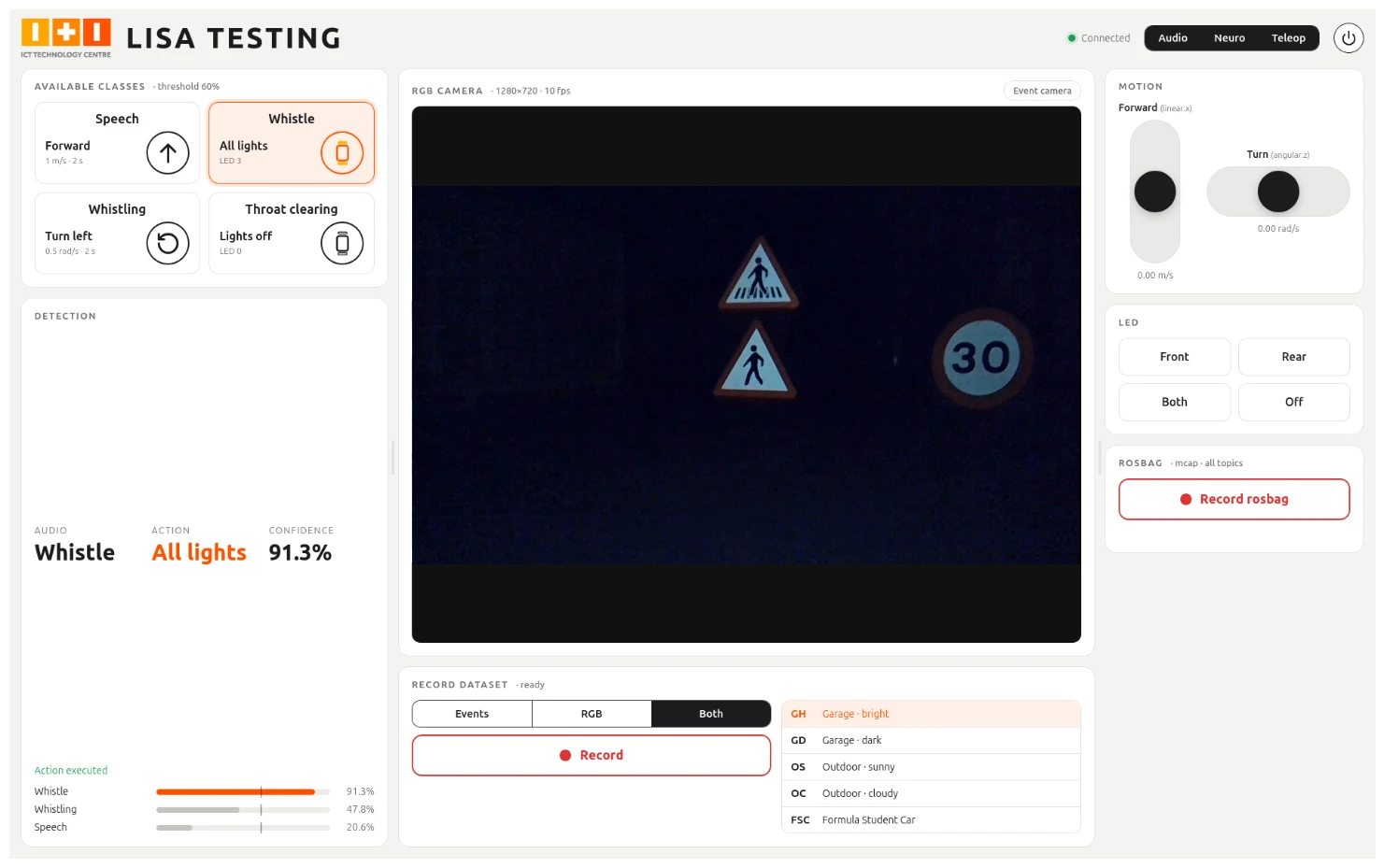}
  \caption{Web dashboard of the robot during operation with the classical audio
  model: a whistle has been detected with a confidence of 91.3\%, triggering the
  action that turns on all the lights.}
  \label{fig:dashboard}
\end{figure}

All software components run inside Docker containers, with one image per target
platform (JetPack~5, 6 and 7, and Raspberry~Pi~4), so that the same software can be
deployed on different devices without modifying the host system. Communication
between machines, for instance between the robot and a laptop used for
teleoperation, relies on the CycloneDDS middleware. In addition, a web dashboard
served from the robot, based on \texttt{rosbridge}, allows any device on the
robot's network to display the audio detections and the actions they trigger, and
to drive the robot and its lights remotely (see Fig.~\ref{fig:dashboard}). The
dashboard shows the classes of the active audio model and their associated
actions, the latest detection with the executed action and the top-3 confidence
values, the live view of the RGB or event camera, and manual controls for motion
and lights. It also allows rosbags and datasets to be recorded from the robot.

\section{System architecture}
\label{sec:architecture}

\begin{figure}[t]
  \centering
  \includegraphics[width=\linewidth]{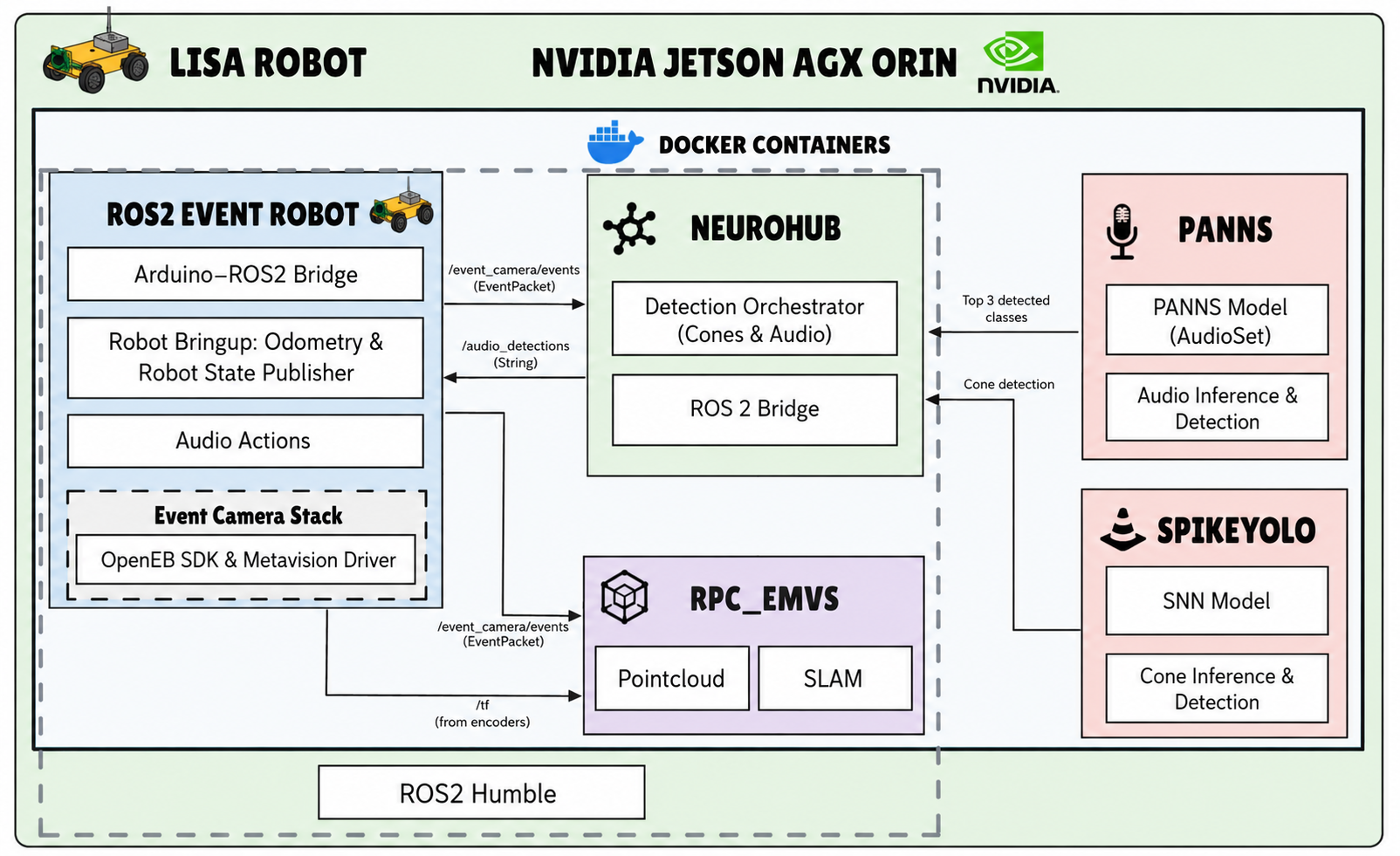}
  \caption{Software architecture of the robot: Docker containers running on the Jetson AGX Orin and ROS~2 topics between them.}
  \label{fig:system-overview}
\end{figure}

The software architecture that integrates the perception pipelines with the robot
is shown in Fig.~\ref{fig:system-overview}. The design follows a modular approach,
in which each perception model runs in its own container and is exposed as an
independent HTTP service, so that models can be developed, replaced or updated
without modifying the rest of the system. Three model services are used in this
work: SpikeYOLO for event-based object detection, PANNs for classical audio tagging
and Xylo for neuromorphic audio classification. SpikSSD~\cite{fan2026spikedetbetterfiringpatterns}, a
spiking detector based on the Single Shot Detector (SSD) architecture, was also
studied as an alternative for the vision pipeline and integrated as a service with
the same interface, but SpikeYOLO was finally selected for the experiments
presented in this report, as it provided the lowest inference time of the two
detectors.

A central orchestrator connects the sensors, the model services and the robot. It
runs two independent pipelines. The vision pipeline is event-driven: events are
read from the ROS~2 event topic, accumulated into fixed time windows, converted
into the representation expected by the detector and sent to the model service by
a dedicated inference worker. The audio pipeline is time-driven: the audio service
is polled at a fixed interval (\SI{250}{ms} by default) to retrieve its latest
predictions. Since the audio models capture the audio themselves, the audio
pipeline does not depend on the event camera and can run with the vision pipeline
disabled. Modules communicate through queues and shared buffers protected by locks,
so that the acquisition of events is never blocked by the inference requests.

The results of both pipelines are published on ROS~2 topics: the bounding boxes on
\path{/detections}, using the standard \path{vision_msgs} format, and the
audio predictions on \path{/audio_detections}, as a JSON list containing the
top-3 classes and their confidence values. On the robot side, an action dispatcher
node subscribes to \path{/audio_detections} and maps the detected classes to
robot behaviours. The dispatcher only implements two primitives: \emph{move},
which publishes a velocity command on \path{/cmd_vel} with a given linear and
angular velocity during a given time, and \emph{led}, which publishes a light
command on \path{/led_robot}. The mapping between classes and actions is not
hardcoded but defined in a YAML file, one per audio model, together with a minimum
confidence threshold (60\% by default). For each message, the dispatcher walks
through the detections from the most likely to the least likely and executes the
action associated with the first class that is present in the mapping and whose
confidence exceeds the threshold. As a result, a new audio model can be added to the
robot simply by providing its action file, without code changes.

\section{Classical audio perception}
\label{sec:panns}

The classical audio pipeline is used as a reference for the neuromorphic one and is
based on the Pretrained Audio Neural Networks (PANNs) framework~\cite{kong2020panns}.
PANNs have become a cornerstone in the field of computational audition, offering
robust and transferable representations across a wide range of audio-related tasks.
These models are trained on AudioSet~\cite{gemmeke2017audio}, a large-scale dataset
of weakly labelled 10-second clips covering more than 500 sound classes, and have
been widely deployed on edge devices for real-time audio
tagging~\cite{grauharo2025comprehensive, naranjoalcazar2025realtime}. In this work,
the CNN14 model trained on AudioSet is used. CNN14 is the PANNs architecture with
the best performance on AudioSet and is widely used in state-of-the-art
applications~\cite{kong2020panns, grauharo2025comprehensive}. Although its
computational cost makes it demanding for CPU-only devices such as the Raspberry
Pi~\cite{grauharo2025comprehensive}, the GPU of the Jetson platforms allows it to
run in real time with a wide margin, as shown below.

\begin{figure}[t]
  \centering
  \includegraphics[width=\linewidth]{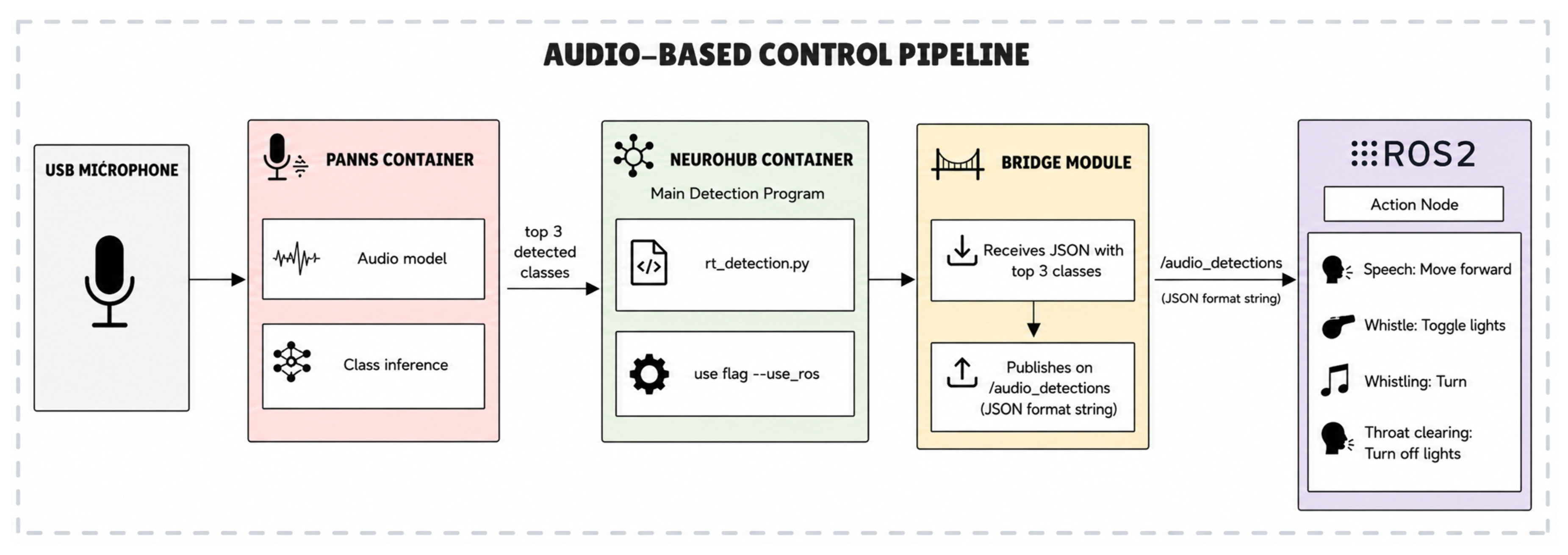}
  \caption{Audio-based control pipeline with the classical audio model, from the
  USB microphone to the actions executed by the robot.}
  \label{fig:panns-pipeline}
\end{figure}

The inference pipeline follows the real-time application presented
in~\cite{grauharo2025comprehensive}, adapted to the Jetson platform. The model is
converted from its original PyTorch checkpoint to the Open Neural Network Exchange
(ONNX) format and executed with ONNX Runtime using the TensorRT execution provider
on the Jetson GPU. The application is composed of independent modules that
communicate through asynchronous queues: an audio recording module captures audio
from the USB microphone, an inference module computes the predictions, and an API
module publishes the latest results so that the orchestrator can retrieve them.
The model used in the experiments expects 10-second inputs sampled at
\SI{32}{kHz}, so that each prediction is computed on a window of 320,000 samples.
This choice aligns with the structure of AudioSet, which is composed of 10-second
clips, providing the model with a sufficient temporal context to capture the
relevant acoustic events. To ensure temporal continuity and reduce the response
time, a sliding window with an 80\% overlap is used, so that a new prediction is
obtained every \SI{2}{s}, each one based on the last \SI{8}{s} of the previous
window and \SI{2}{s} of new audio. In addition, versions of the model with
shorter analysis windows, of \SI{4}{s} and \SI{2}{s}, are available and can be
selected depending on the response time required by the application. The three classes with the highest scores are returned together with
their confidence values. The complete audio-based control pipeline, from the
microphone to the robot actions, is shown in Fig.~\ref{fig:panns-pipeline}.

These configurations are feasible because the inference time of the model is short
compared with the time between predictions. Table~\ref{tab:panns-times} reports the
mean inference time of CNN14 on the two Jetson platforms considered in this work,
with the board in its maximum performance power mode, when the model is executed on
the CPU, on the GPU with CUDA and on the GPU with TensorRT. The CUDA and TensorRT
times were averaged over 24 hours of continuous inference, while the CPU times were
averaged over 100 consecutive windows. With TensorRT, the mean inference time is
\SI{8.4}{ms} on the AGX Orin and \SI{25.0}{ms} on the Orin Nano, two orders of
magnitude below the \SI{2}{s} interval between predictions of the 10-second
configuration.

\begin{table}[t]
  \centering
  \caption{Mean inference time of CNN14 on the Jetson platforms (10-second input).}
  \label{tab:panns-times}
  \begin{tabular}{lccc}
    \toprule
    Platform (power mode) & CPU & CUDA & TensorRT FP32 \\
    \midrule
    Jetson AGX Orin (MAXN) & \SI{165.5}{ms} & \SI{48.3}{ms} & \SI{8.4}{ms} \\
    Jetson Orin Nano (MAXN SUPER) & \SI{355.8}{ms} & \SI{75.7}{ms} & \SI{25.0}{ms} \\
    \bottomrule
  \end{tabular}
\end{table}

As AudioSet-trained models may behave unexpectedly when deployed in real
environments~\cite{garciaballesteros2026confidently}, the classes used to control
the robot were selected experimentally: among the AudioSet classes that can be
easily produced in front of the robot, those that were detected most reliably and
consistently in our experiments were chosen. Table~\ref{tab:actions} summarises the
actions associated with these classes.

\begin{table}[t]
  \centering
  \caption{Mapping between audio classes and robot actions for each audio model.}
  \label{tab:actions}
  \begin{tabular}{lll}
    \toprule
    Model & Class & Action \\
    \midrule
    PANNs & Speech & Move forward (\SI{1.0}{m/s}, \SI{2}{s}) \\
          & Whistle & Turn on all lights \\
          & Whistling & Rotate in place (\SI{0.5}{rad/s}, \SI{2}{s}) \\
          & Throat clearing & Turn off all lights \\
    \midrule
    Xylo  & High-pitch & Move forward (\SI{1.0}{m/s}, \SI{2}{s}) \\
          & Low-pitch & Move backward (\SI{1.0}{m/s}, \SI{2}{s}) \\
          & Silence & Stop \\
    \bottomrule
  \end{tabular}
\end{table}

\section{Neuromorphic audio perception on Xylo}
\label{sec:xylo}

This section describes the neuromorphic audio pipeline of the robot. First, the
Xylo Audio~2 platform is introduced. Next, the classification task and the SNN
architecture are presented, followed by the training procedure. Then, the
deployment of the trained network on the chip and the decoding of its outputs are
detailed, together with the integration of the pipeline in the robot.

\subsection{The Xylo Audio~2 platform}
\label{sec:xylo-platform}

Xylo Audio~2 (SYNS61201) is a neuromorphic processor developed by SynSense for
low-power audio processing at the edge~\cite{bos2022sub}. The chip integrates two
main blocks. The first one is an analog audio front-end (AFE), which amplifies the
microphone signal and splits it into 16 frequency channels by means of a bank of
band-pass filters. The output of each filter is rectified and converted into spike
trains, so that the audio signal is encoded into 16 spiking input channels without
any processing on the host. The second block is a digital SNN core that executes
networks of LIF neurons. 


In the digital core, the synaptic weights are
represented as integers and the exponential decays are implemented as bit-shift
operations, so that the time constants are quantised to the values that can be
represented by the hardware. 


Xylo Audio~2 is used through its hardware development kit, which connects to the
host via USB and is controlled with the Rockpool~\cite{muir2019rockpool} and Samna
libraries. Rockpool provides a complete workflow to train SNNs in PyTorch, simulate
the AFE and the digital core, map the trained networks to the hardware resources,
and deploy and run them on the chip.

\subsection{Task definition}

The neuromorphic audio pipeline addresses the classification of sounds into two
classes, \emph{low-pitch} and \emph{high-pitch}, according to their dominant
frequency content. In addition, an implicit \emph{silence} class is considered when
none of the output neurons is active, so that the robot does not react in the
absence of sound. This task was chosen because it can be solved by a very small
network, its classes can be easily produced during a demonstration (for instance,
by whistling or humming at different pitches) and it is naturally suited to the
frequency decomposition performed by the AFE. 


\subsection{Network architecture}

\begin{figure}[t]
  \centering
  \begin{tikzpicture}[x=1cm, y=1cm]
  \foreach \i/\lab in {1/1, 2/2, 3/3, 5/14, 6/15, 7/16} {
    \node[neuron, fill=chipfill] (in\i) at (0, {3.5-\i}) {\lab};
  }
  \node at (0, -0.5) {$\vdots$};
  \node[neuron, fill=hostfill] (h1) at (3.5, 1.2) {LIF};
  \node[neuron, fill=hostfill] (h2) at (3.5, -2.2) {LIF};
  \node[neuron, fill=robotfill] (o1) at (7, 1.2) {LIF};
  \node[neuron, fill=robotfill] (o2) at (7, -2.2) {LIF};
  \node[right=3mm of o1, font=\footnotesize, align=left] {High-pitch};
  \node[right=3mm of o2, font=\footnotesize, align=left] {Low-pitch};
  \begin{scope}[on background layer]
    \foreach \i in {1,2,3,5,6,7} {
      \foreach \h in {h1,h2} { \draw[gray!70] (in\i) -- (\h); }
    }
    \foreach \h in {h1,h2} {
      \foreach \o in {o1,o2} { \draw[gray!70] (\h) -- (\o); }
    }
  \end{scope}
  \draw[arr] (-1.1, 2.5) -- (-1.1, -3.5)
    node[midway, rotate=90, above, font=\scriptsize] {centre frequency};
  \node[font=\footnotesize, align=center] at (0, -4.5)
    {AFE channels\\(16 inputs)};
  \node[font=\footnotesize, align=center] at (3.5, -4.5)
    {Hidden layer\\(2 LIF neurons)};
  \node[font=\footnotesize, align=center] at (7, -4.5)
    {Output layer\\(2 LIF neurons)};
  \node[font=\scriptsize] at (1.75, 2.3) {$W_\text{in} \in \mathbb{R}^{16\times2}$};
  \node[font=\scriptsize] at (5.25, 2.3) {$W_\text{out} \in \mathbb{R}^{2\times2}$};
\end{tikzpicture}
  \caption{Architecture of the spiking neural network deployed on Xylo Audio~2.}
  \label{fig:xylo-snn}
\end{figure}
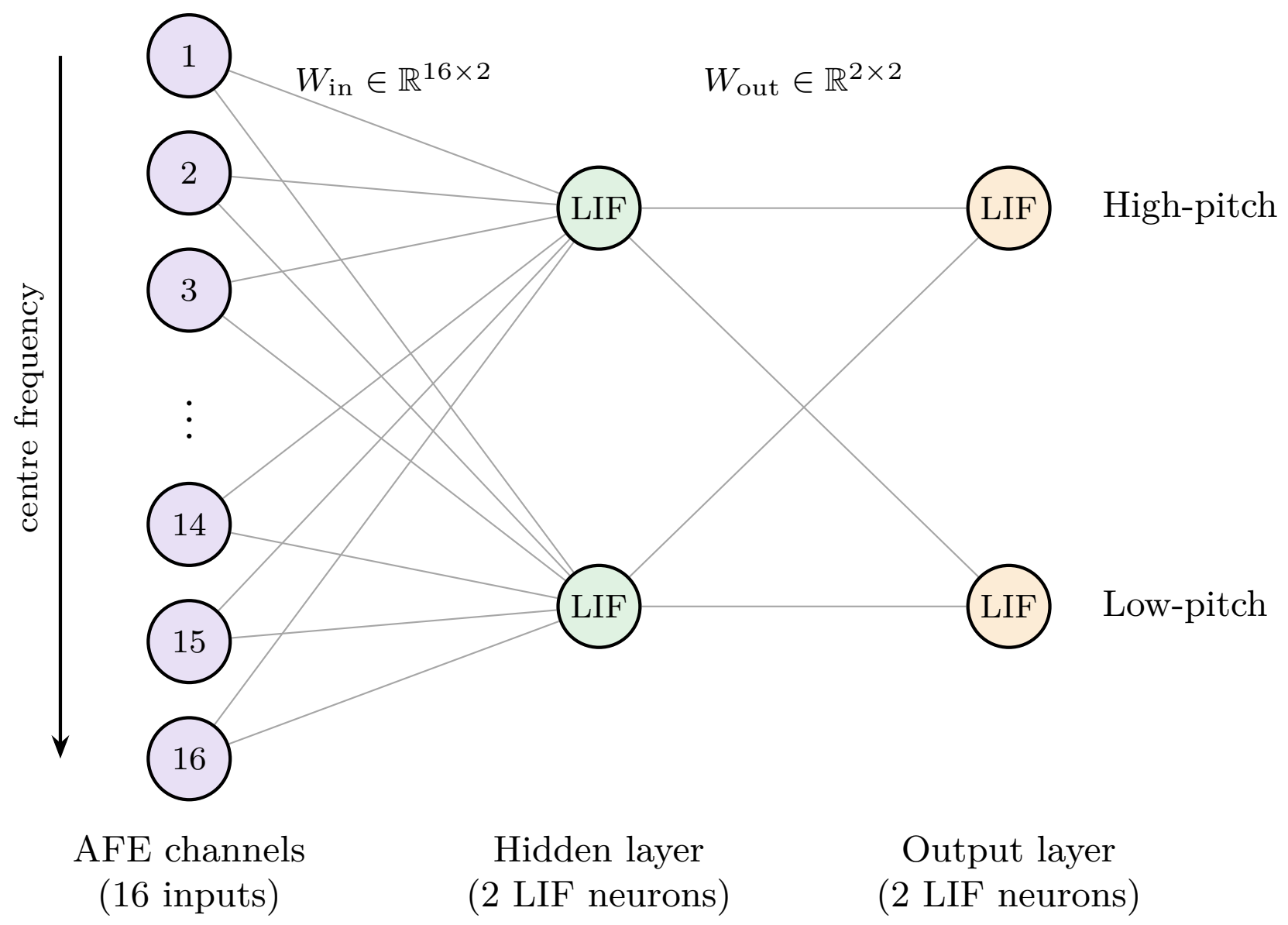

The SNN deployed on the chip is based on the \texttt{SynNet} architecture provided by
Rockpool~\cite{muir2019rockpool, bos2022sub}, a feed-forward network of LIF neurons
specifically designed for audio classification on Xylo. The network is shown in
Fig.~\ref{fig:xylo-snn} and its hyperparameters are summarised in
Table~\ref{tab:xylo-params}. The input layer has 16 channels, one per AFE
filter. The network has a single hidden layer of only two LIF neurons with a single
synaptic time constant, followed by an output layer of two LIF neurons, one per
class. All neurons use a synaptic and a membrane time constant of \SI{20}{ms}, a
threshold of 1.0 and a time step of \SI{10}{ms}, which matches the rate at which the
AFE delivers spikes to the digital core. The time constants and thresholds are kept
fixed during training, and the number of spikes that a neuron can emit per time
step is limited to 15 in the hidden layer, which is the maximum supported by the
hardware, and to one in the output layer.

The resulting network has only four neurons and 


trainable parameters, which uses
a negligible fraction of the resources of the chip. Such a small network is
sufficient for this task because the AFE already performs a frequency decomposition
of the signal: distinguishing between low- and high-pitched sounds mainly requires
weighting the activity of the low- and high-frequency channels, which can be
achieved with very few neurons. 


\begin{table}[t]
  \centering
  \caption{Hyperparameters of the SNN deployed on Xylo Audio~2.}
  \label{tab:xylo-params}
  \begin{tabular}{ll}
    \toprule
    Parameter & Value \\
    \midrule
    Architecture & Rockpool \texttt{SynNet} \\
    Neuron model & LIF (\texttt{LIFTorch}) \\
    Input channels & 16 (AFE filter bank) \\
    Hidden layers & 1 layer of 2 neurons \\
    Synaptic time constants per layer & 1 \\
    Output neurons & 2 \\
    $\tau_\text{syn}$, $\tau_\text{mem}$ & \SI{20}{ms} \\
    Threshold & 1.0 (not trained) \\
    Time step $dt$ & \SI{10}{ms} \\
    Max.\ spikes per time step & 15 (hidden), 1 (output) \\
    Learning rate & 0.01 \\
    Epochs & 30 \\
    \bottomrule
  \end{tabular}
\end{table}



\subsection{Deployment on the chip}

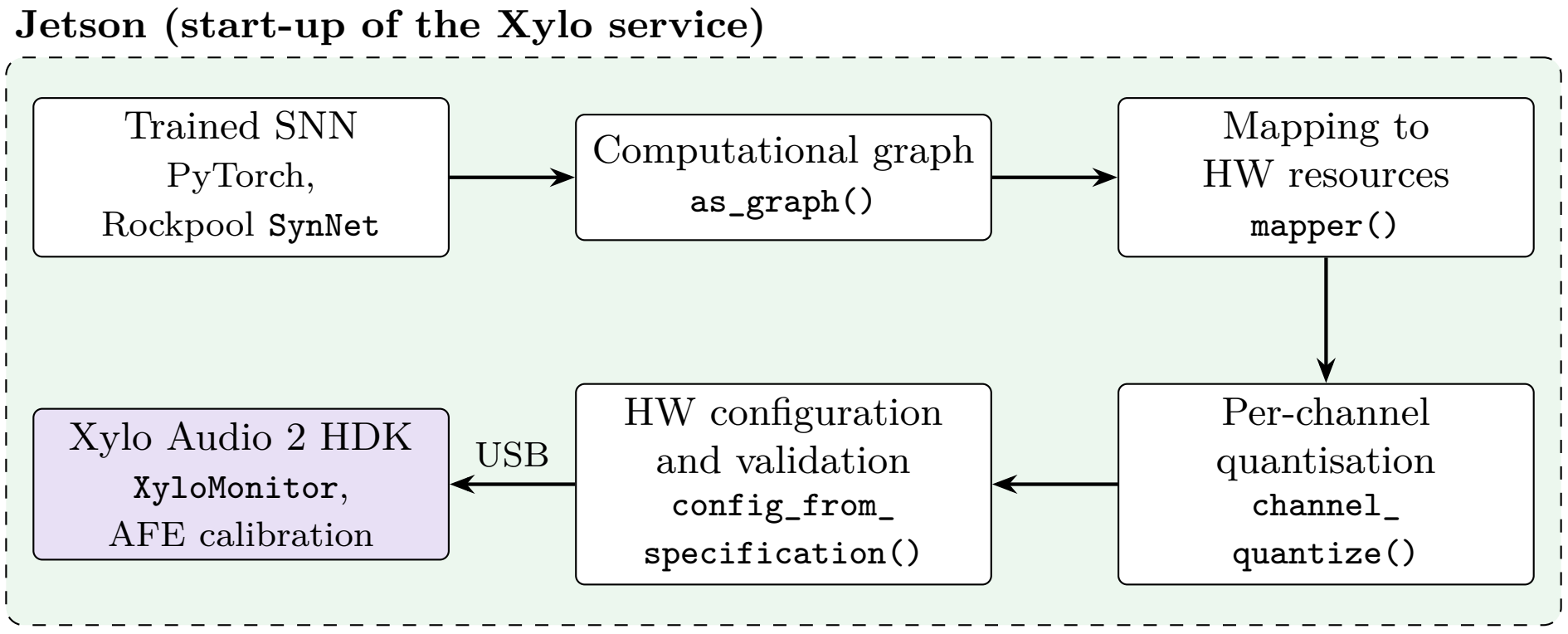
\begin{figure}[t]
  \centering
  \begin{tikzpicture}[node distance=5mm and 10mm]
  \node[blk, text width=3.1cm] (net) {Trained SNN\\\scriptsize PyTorch, Rockpool \texttt{SynNet}};
  \node[blk, text width=3.1cm, right=of net] (graph) {Computational graph\\\scriptsize \texttt{as\_graph()}};
  \node[blk, text width=3.1cm, right=of graph] (map) {Mapping to HW resources\\\scriptsize \texttt{mapper()}};
  \node[blk, text width=3.1cm, below=10mm of map] (quant) {Per-channel quantisation\\\scriptsize \texttt{channel\_}\\\texttt{quantize()}};
  \node[blk, text width=3.1cm, left=of quant] (conf) {HW configuration and validation\\\scriptsize \texttt{config\_from\_}\\\texttt{specification()}};
  \node[blk, text width=3.1cm, left=of conf, fill=chipfill] (hdk) {Xylo Audio~2 HDK\\\scriptsize \texttt{XyloMonitor}, AFE calibration};
  \draw[arr] (net) -- (graph);
  \draw[arr] (graph) -- (map);
  \draw[arr] (map) -- (quant);
  \draw[arr] (quant) -- (conf);
  \draw[arr] (conf) -- node[above, font=\scriptsize] {USB} (hdk);
  \begin{scope}[on background layer]
    \node[grp, fill=hostfill!60, fit=(net)(graph)(map)(quant)(conf),
      inner ysep=9pt] (host) {};
  \end{scope}
  \node[grplabel] at (host.north west) {Jetson (start-up of the Xylo service)};
\end{tikzpicture}
  \caption{Deployment flow from the trained model to the Xylo Audio~2 chip.}
  \label{fig:xylo-deploy}
\end{figure}

The deployment of the trained network on Xylo Audio~2 follows the workflow provided
by Rockpool and is illustrated in Fig.~\ref{fig:xylo-deploy}. First, the trained
PyTorch network is loaded and converted into a computational graph that describes
its neurons, connections and parameters independently of the training framework.
Next, this graph is mapped to the hardware resources of the chip, assigning each
neuron of the network to a physical neuron of the digital core. The floating-point
weights, thresholds and time constants are then quantised to the integer
representation supported by the hardware, using a per-channel quantisation scheme.
From the quantised specification, a hardware configuration is generated and
validated, ensuring that the network satisfies all the constraints of the chip.

The configuration is finally written to the chip through the development kit. At
this stage, the AFE is configured with a high amplification level and its
auto-calibration is enabled, a process that takes approximately \SI{45}{s} after the
board is initialised. Neither divisive normalisation nor the hibernation mode of
the chip are used. 


Once the network is deployed, the audio signal never leaves the chip during
inference: the microphone signal is converted into spikes by the AFE and processed
by the SNN core, while the host only triggers the evolution of the chip and reads
the state of the output neurons. This is a key difference with respect to the
classical pipeline, in which the raw audio has to be captured, transferred and
processed by the Jetson.

\subsection{Output decoding and integration in the robot}

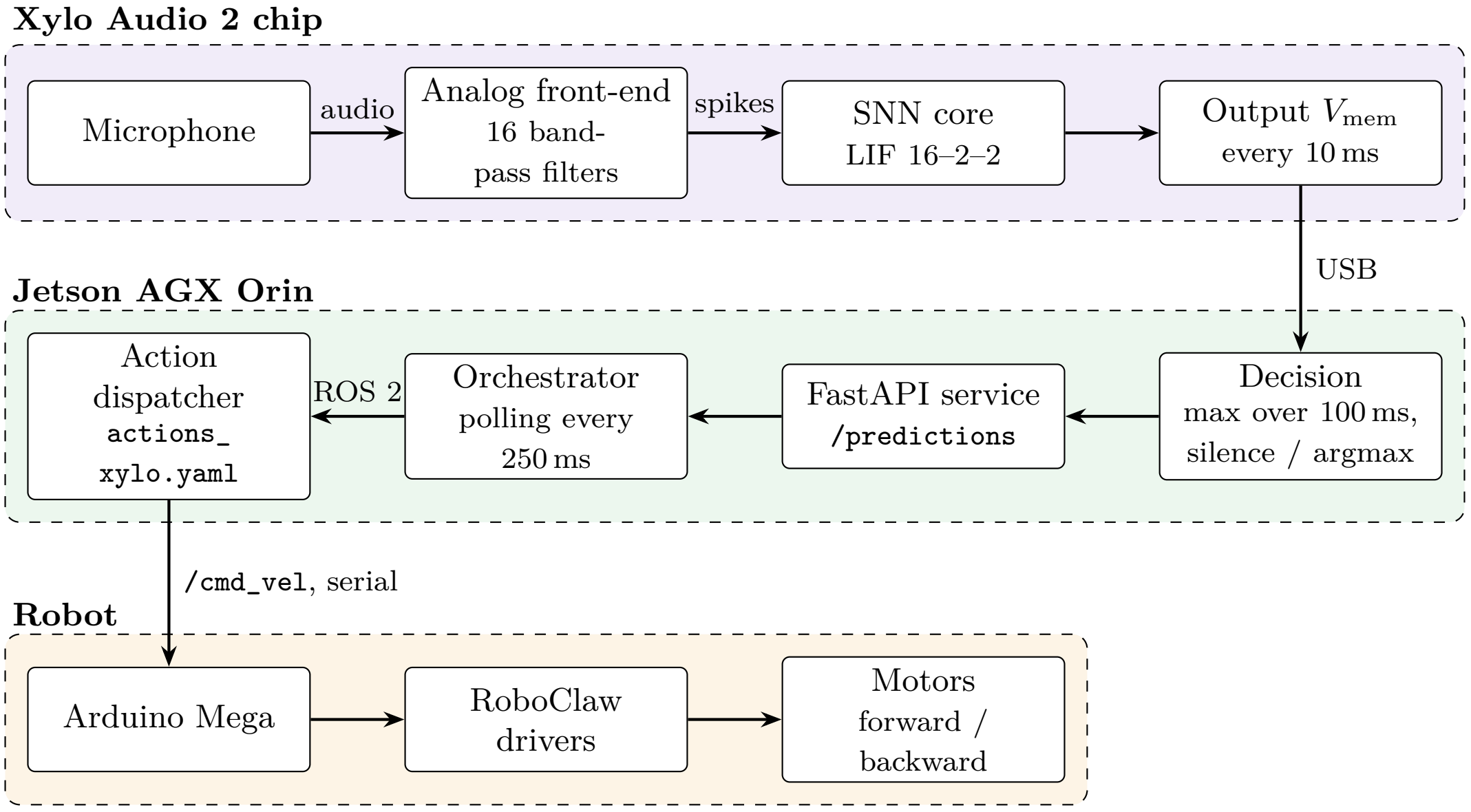
\begin{figure}[t]
  \centering
  \begin{tikzpicture}[node distance=4mm and 9mm]
  \node[blk, text width=2.5cm] (mic) {Microphone};
  \node[blk, text width=2.5cm, right=of mic] (afe) {Analog front-end\\\scriptsize 16 band-pass filters};
  \node[blk, text width=2.5cm, right=of afe] (snn) {SNN core\\\scriptsize LIF 16--2--2};
  \node[blk, text width=2.5cm, right=of snn] (vmem) {Output $V_\text{mem}$\\\scriptsize every \SI{10}{ms}};
  \draw[arr] (mic) -- node[above, font=\scriptsize] {audio} (afe);
  \draw[arr] (afe) -- node[above, font=\scriptsize] {spikes} (snn);
  \draw[arr] (snn) -- (vmem);
  \node[blk, text width=2.5cm, below=16mm of vmem] (dec) {Decision\\\scriptsize max over \SI{100}{ms},\\silence / argmax};
  \node[blk, text width=2.5cm, left=of dec] (api) {FastAPI service\\\scriptsize \texttt{/predictions}};
  \node[blk, text width=2.5cm, left=of api] (orch) {Orchestrator\\\scriptsize polling every \SI{250}{ms}};
  \node[blk, text width=2.5cm, left=of orch] (disp) {Action dispatcher\\\scriptsize \texttt{actions\_}\\\texttt{xylo.yaml}};
  \draw[arr] (vmem) -- node[right, font=\scriptsize] {USB} (dec);
  \draw[arr] (dec) -- (api);
  \draw[arr] (api) -- (orch);
  \draw[arr] (orch) -- node[above, font=\scriptsize] {ROS~2} (disp);
  \node[blk, text width=2.5cm, below=16mm of disp] (ard) {Arduino Mega};
  \node[blk, text width=2.5cm, right=of ard] (rc) {RoboClaw drivers};
  \node[blk, text width=2.5cm, right=of rc] (mot) {Motors\\\scriptsize forward / backward};
  \draw[arr] (disp) -- node[right, font=\scriptsize] {\texttt{/cmd\_vel}, serial} (ard);
  \draw[arr] (ard) -- (rc);
  \draw[arr] (rc) -- (mot);
  \begin{scope}[on background layer]
    \node[grp, fill=chipfill!60, fit=(mic)(afe)(snn)(vmem)] (g1) {};
    \node[grp, fill=hostfill!60, fit=(disp)(orch)(api)(dec)] (g2) {};
    \node[grp, fill=robotfill!60, fit=(ard)(rc)(mot)] (g3) {};
  \end{scope}
  \node[grplabel] at (g1.north west) {Xylo Audio~2 chip};
  \node[grplabel] at (g2.north west) {Jetson AGX Orin};
  \node[grplabel] at (g3.north west) {Robot};
\end{tikzpicture}
  \caption{End-to-end neuromorphic audio pipeline, from sound to robot motion.}
  \label{fig:xylo-pipeline}
\end{figure}

The end-to-end neuromorphic audio pipeline is shown in Fig.~\ref{fig:xylo-pipeline}.
During inference, the chip is evolved in blocks of 10 time steps, that is,
\SI{100}{ms} of audio, and the membrane potentials $V_\text{mem}$ of the two
output neurons are read at each block. For each output neuron, the maximum membrane
potential within the block is computed. If both values are zero, the block is
classified as \emph{silence}; otherwise, the class associated with the output
neuron with the highest value is selected. As a result, the neuromorphic pipeline
produces a new prediction every \SI{100}{ms}, compared to the \SI{2}{s} of the
classical pipeline and its 10-second analysis window.

The inference loop runs in a dedicated thread inside the Xylo container, which
stores each new prediction in a queue. A second thread consumes this queue and
updates the state of an HTTP API, built with FastAPI, that exposes the latest
prediction. The orchestrator polls this API as described in
Section~\ref{sec:architecture} and publishes the result on
\path{/audio_detections}, where it is received by the action dispatcher. Using
the action file of the Xylo model (see Table~\ref{tab:actions}), a high-pitched
sound makes the robot move forward for \SI{2}{s}, a low-pitched sound makes it move
backward for \SI{2}{s}, and silence produces no movement. Since the current decoding
is based on the winning output neuron, the service reports a fixed confidence value
of 100\% for each prediction.

\section{Neuromorphic vision perception}
\label{sec:vision}

\subsection{Event-based sensing and representation}

The event camera is managed by the robot through its ROS~2 driver, which publishes
the event stream on \path{/event_camera/events}. The orchestrator subscribes to
this topic, decodes the events and accumulates them into consecutive time windows of
\SI{250}{ms}.

Since the detector operates on dense tensors, the events of each window are
converted into a frame-like representation. The representation is initialised to
a neutral grey value of 127, and each pixel that receives an event is set to 255 or
0 according to the polarity of its most recent event, so that positive and negative
brightness changes appear as bright and dark pixels over a grey background. The
representation is built at the resolution of the sensor and replicated in three
channels to match the input format of the detector. All the events of each
\SI{250}{ms} window are collapsed into a single frame.

\subsection{SpikeYOLO detector}

Object detection is performed with SpikeYOLO~\cite{luo2024integer}, a spiking
object detector that simplifies the YOLOv8 architecture and incorporates meta SNN
blocks. SpikeYOLO introduces the Integer LIF (I-LIF) neuron, which is trained with
integer-valued activations to reduce quantisation errors and is converted into
binary spikes at inference time, preserving the spike-driven nature of the
computation. SpikeYOLO achieves state-of-the-art performance among SNNs on the COCO
and Gen1~\cite{detournemire2020large} datasets, the latter being an automotive
detection dataset recorded with an event camera.

In this work, only the SpikeYOLO architecture is reused: the network was adapted
to the HD resolution of the event camera of the robot ($1280\times720$ pixels) and
trained from scratch for cone detection on event data recorded by the research
group with the same Prophesee EVK3 HD camera, so that the training data matches the
sensor and the representation used at inference time. No pretrained weights from
the original work are used.

\subsection{Integration in the robot}

The detector runs in its own container on the Jetson GPU and exposes an HTTP
endpoint that receives a voxel grid and returns the detected bounding boxes. In the
orchestrator, the construction of the voxel grids and the inference requests are
decoupled by a queue, so that a dedicated inference worker sends each voxel grid to
the service while new events continue to be accumulated. The detections whose
confidence exceeds a threshold are published on \path{/detections} and can be
visualised in RViz together with the robot model. In the current version of the
system, the vision pipeline is used for perception only: the detected cones are
published and visualised, but they are not yet mapped to robot actions.

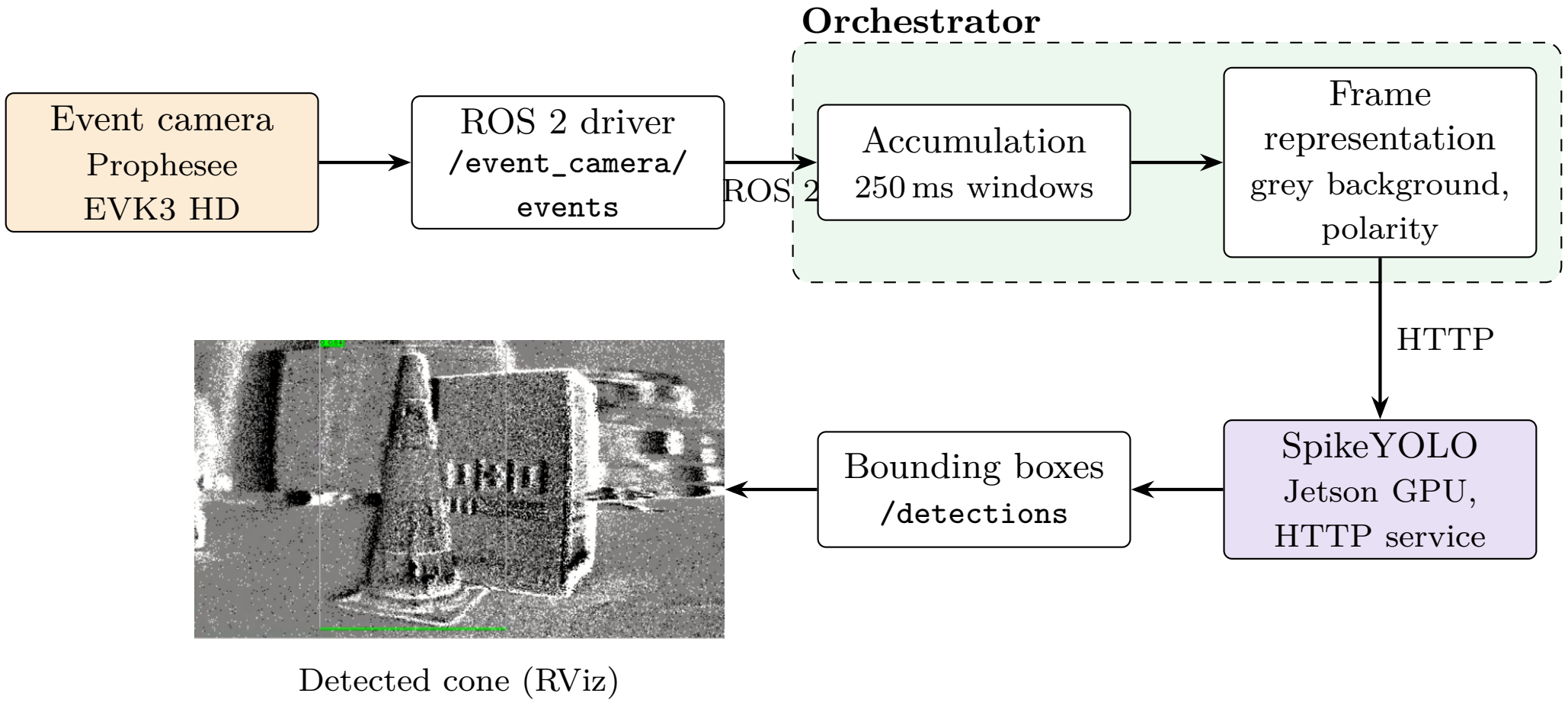
\begin{figure}[t]
  \centering
  \begin{tikzpicture}[node distance=4mm and 8mm]
  \node[blk, text width=2.5cm, fill=robotfill] (cam) {Event camera\\\scriptsize Prophesee EVK3 HD};
  \node[blk, text width=2.5cm, right=of cam] (drv) {ROS~2 driver\\\scriptsize \texttt{/event\_camera/}\\\texttt{events}};
  \node[blk, text width=2.5cm, right=of drv] (acc) {Accumulation\\\scriptsize \SI{250}{ms} windows};
  \node[blk, text width=2.5cm, right=of acc] (frm) {Frame representation\\\scriptsize grey background, polarity};
  \node[blk, text width=2.5cm, below=14mm of frm, fill=chipfill] (yolo) {SpikeYOLO\\\scriptsize Jetson GPU, HTTP service};
  \node[blk, text width=2.5cm, left=of yolo] (det) {Bounding boxes\\\scriptsize \texttt{/detections}};
  \node[left=8mm of det, inner sep=0pt] (img)
    {\includegraphics[width=4.6cm]{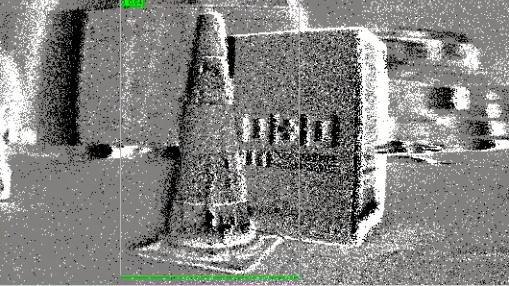}};
  \node[below=1mm of img, font=\scriptsize] {Detected cone (RViz)};
  \draw[arr] (cam) -- (drv);
  \draw[arr] (drv) -- node[below, font=\scriptsize] {ROS~2} (acc);
  \draw[arr] (acc) -- (frm);
  \draw[arr] (frm) -- node[right, font=\scriptsize] {HTTP} (yolo);
  \draw[arr] (yolo) -- (det);
  \draw[arr] (det) -- (img);
  \begin{scope}[on background layer]
    \node[grp, fill=hostfill!60, fit=(acc)(frm)] (g1) {};
  \end{scope}
  \node[grplabel] at (g1.north west) {Orchestrator};
\end{tikzpicture}
  \caption{Event-based vision pipeline, from the event camera to the cone detections published in ROS~2. The image shows a real detection over the frame representation built from the events, visualised in RViz.}
  \label{fig:vision}
\end{figure}

\section{Proof-of-concept demonstration}
\label{sec:evaluation}

The aim of this report is not to provide a quantitative benchmark of the
perception models, but to demonstrate that classical and neuromorphic perception
pipelines can be integrated and run together on the same robot, and that their
outputs can drive the robot in real time. This section describes the
proof-of-concept demonstration carried out with the platform and discusses,
qualitatively, the differences between the two audio pipelines.

Each perception capability was demonstrated separately, in an environment chosen
to suit the characteristics of the corresponding sensor and task. The three
scenarios are described below.

\subsection{Neuromorphic audio}

The neuromorphic audio pipeline was tested in a semi-anechoic room, in the absence
of background noise. This controlled environment ensured that the activity of the
output neurons of the chip was caused only by the sounds produced during the test,
which is particularly relevant given the simplicity of the task and of the
deployed network. The operators produced low- and high-pitched sounds in front of
the robot, and the robot executed the corresponding action: high-pitched sounds
made it move forward and low-pitched sounds made it move backward, while it
remained still in the absence of sound. During the test, the whole audio
processing took place on the Xylo chip, and the Jetson only read the output
neurons and forwarded the result to the robot.

\subsection{Classical audio}

The classical audio pipeline was tested in a similar way, but without the need for
a semi-anechoic room, since CNN14 is trained on AudioSet, which covers a wide variety
of real-world recording conditions~\cite{gemmeke2017audio}. The operators produced
the sound of the class associated with the desired action, and the robot reacted
according to the mapping of Table~\ref{tab:actions}: speaking made it move forward,
whistling made it rotate in place, and whistles and throat clearing turned its
lights on and off, respectively. The detections and the executed actions were
monitored in real time through the web dashboard (Fig.~\ref{fig:dashboard}).

\subsection{Event-based cone detection}

For the vision pipeline, a circuit with traffic cones was set up for the robot to
detect along its way, and the robot was teleoperated through the circuit with a
gamepad. Since event cameras respond to changes in brightness, the
ambient lighting can generate a large number of background events that are not
related to the objects of interest. To reduce this event noise, the test was
carried out with the lights of the room switched off, using only the LEDs of the
robot to illuminate the scene. Under these conditions, SpikeYOLO detected the cones
of the circuit, and the detections were published on \path{/detections} and
visualised in RViz together with the robot model. An example of a cone detected
during the test is shown in Fig.~\ref{fig:vision}.

\subsection{Classical vs.\ neuromorphic audio pipelines}

Table~\ref{tab:audio-comparison} summarises the main differences between the two
audio pipelines, as derived from their design. The classical pipeline offers a
rich vocabulary of sound classes, inherited from AudioSet, at the cost of capturing
and processing the raw audio on the Jetson GPU with a large CNN and long analysis
windows. The neuromorphic pipeline solves a much simpler task with a network of
only four neurons, but it does so entirely on the chip, producing a new prediction
every \SI{100}{ms} without any audio processing on the host. Both approaches are
therefore complementary: the former is suited to recognising a wide range of sound
events, while the latter is suited to simple, always-on audio triggers with a
minimal load on the main computer. A quantitative comparison in terms of latency,
energy consumption and accuracy is left for future work.

\begin{table}[t]
  \centering
  \caption{Main characteristics of the classical and the neuromorphic audio
  pipelines.}
  \label{tab:audio-comparison}
  \begin{tabular}{lll}
    \toprule
    & Classical (PANNs) & Neuromorphic (Xylo) \\
    \midrule
    Hardware & Jetson GPU & Xylo Audio~2 \\
    Model & CNN14 & \texttt{SynNet} (4 neurons) \\
    Number of classes & 527 (AudioSet) & 2 + silence \\
    Analysis window & \SI{10}{s} (also \SI{4}{s}, \SI{2}{s}) & \SI{100}{ms} \\
    Prediction interval & \SI{2}{s} & \SI{100}{ms} \\
    Audio processed by the host & Yes & No \\
    \bottomrule
  \end{tabular}
\end{table}

\section{Limitations and future work}
\label{sec:limitations}

The current system presents several limitations. First, the neuromorphic audio
task is restricted to a binary low- versus high-pitch classification, which
demonstrates the integration of the Xylo chip in the robot but is far from the
variety of sound classes handled by the classical pipeline. Second, the Xylo
service reports the winning class without a calibrated confidence value. Third, the
detections of the vision pipeline are published and visualised but do not yet
trigger robot behaviours. Finally, the system has been validated through a
proof-of-concept demonstration, and a systematic quantitative evaluation has not
yet been carried out.

As future work, the neuromorphic audio pipeline will be extended to richer sets of
sound classes, taking advantage of the remaining resources of the chip and of
efficient spike encoding schemes~\cite{larroza2025spike}. In addition,
vision-driven behaviours, such as stopping or avoiding obstacles when cones are
detected, will be added to the action dispatcher, and the fusion of audio and visual
information will be explored. Finally, a systematic experimental evaluation of
the platform will be conducted, including the end-to-end latency of each pipeline,
the accuracy of the models in the robot's operating conditions and the energy
consumption of the classical and neuromorphic pipelines, in order to quantify the
benefits of neuromorphic processing at system level.



\section*{Acknowledgements}


This work was partially funded by the Valencian Institute for Business Competitiveness (IVACE) and FEDER funds through the SAVANt project (IMDEEA/2025/88). This work has been supported by Grants \mbox{PID2025-174060OB-C31 and PID2022-137048OA-C44 } funded by MICIU/AEI/10.13039/501100011033 and by ``ERDF/EU".




\bibliographystyle{unsrt}
\bibliography{references}

\end{document}